\documentclass[journal, web]{ieeecolor}
\usepackage{tmi}
\usepackage{cite}
\usepackage{amsmath,amssymb,amsfonts}
\usepackage{algorithmic}
\usepackage{graphicx}
\usepackage{textcomp}
\usepackage{multirow}
\def\BibTeX{{\rm B\kern-.05em{\sc i\kern-.025em b}\kern-.08em
    T\kern-.1667em\lower.7ex\hbox{E}\kern-.125emX}}
\begin{document}
\title{Adapting 2D Multi-Modal Large Language Model for 3D CT Image Analysis}
\author{Yang Yu, Dunyuan Xu, Yaoqian Li, Xiaomeng Li \IEEEmembership{Senior Member, IEEE}, Jinpeng Li, and Pheng-Ann Heng \IEEEmembership{Senior Member, IEEE}
\thanks{Yang Yu and Xiaomeng Li are with Department of Electronic and Computer Engineering, The Hong Kong University of Science and Technology, Hong Kong, China. (email: eeyangyu@ust.hk)}
\thanks{Dunyuan Xu, Yaoqian Li, and Pheng-Ann Heng are with Department of Computer Science and Engineering, The Chinese University of Hong Kong, Hong Kong, China.}
\thanks{Pheng-Ann Heng is also with the Institute of Medical Intelligence and XR, The Chinese University of Hong Kong, Hong Kong, China.} 
\thanks{Jinpeng Li is with Center for Artificial Intelligence and Robotics, Hong Kong Institute of Science and Innovation, Chinese Academy of Sciences, Hong Kong, China. (email: jinpeng.li@cair-cas.org.hk)}
\thanks{Corresponding author: Jinpeng Li}
}

\maketitle

\begin{abstract}
3D medical image analysis is of great importance in disease diagnosis and treatment.
Recently, multimodal large language models (MLLMs) have exhibited robust perceptual capacity, strong cross-modal alignment, and promising generalizability. Therefore, they have great potential to improve the performance of medical report generation (MRG) and medical visual question answering (MVQA), which serve as two important tasks in clinical scenarios. However, due to the scarcity of 3D medical images, existing 3D medical MLLMs suffer from insufficiently pretrained vision encoder and inability to extract customized image features for different kinds of tasks. In this paper, we propose to first transfer a 2D MLLM, which is well trained with 2D natural images, to support 3D medical volumetric inputs while reusing all of its pre-trained parameters. To enable the vision encoder to extract tailored image features for various tasks, we then design a Text-Guided Hierarchical MoE (TGH-MoE) framework, which can distinguish tasks under the guidance of the text prompt.
Furthermore, we propose a two-stage training strategy to learn both task-shared and task-specific image features.  As demonstrated empirically, our method outperforms existing 3D medical MLLMs in both MRG and MVQA tasks. Our code will be released once this paper is accepted.
\end{abstract}


\section{Introduction}
\label{sec:introduction}
\IEEEPARstart{M}{ultimodal} Large Language Models 
\cite{qwen2025qwen25technicalreport ,chen2024internvl}, which exhibit strong multimodal ability, are expected to be assistants to radiologists and have attracted increasing research interest. 
However, most MLLMs \cite{abdin2024phi3technicalreporthighly ,chen2024internvl} are pretrained on massive natural images and text which exhibit a large domain gap with 3D medical images and text. Bridging the gap between MLLMs trained on 2D natural images and 3D medical volumetric scans is a crucial challenge in the intelligent healthcare domain.

\begin{figure}[!t]
\centerline{\includegraphics[width=\columnwidth]{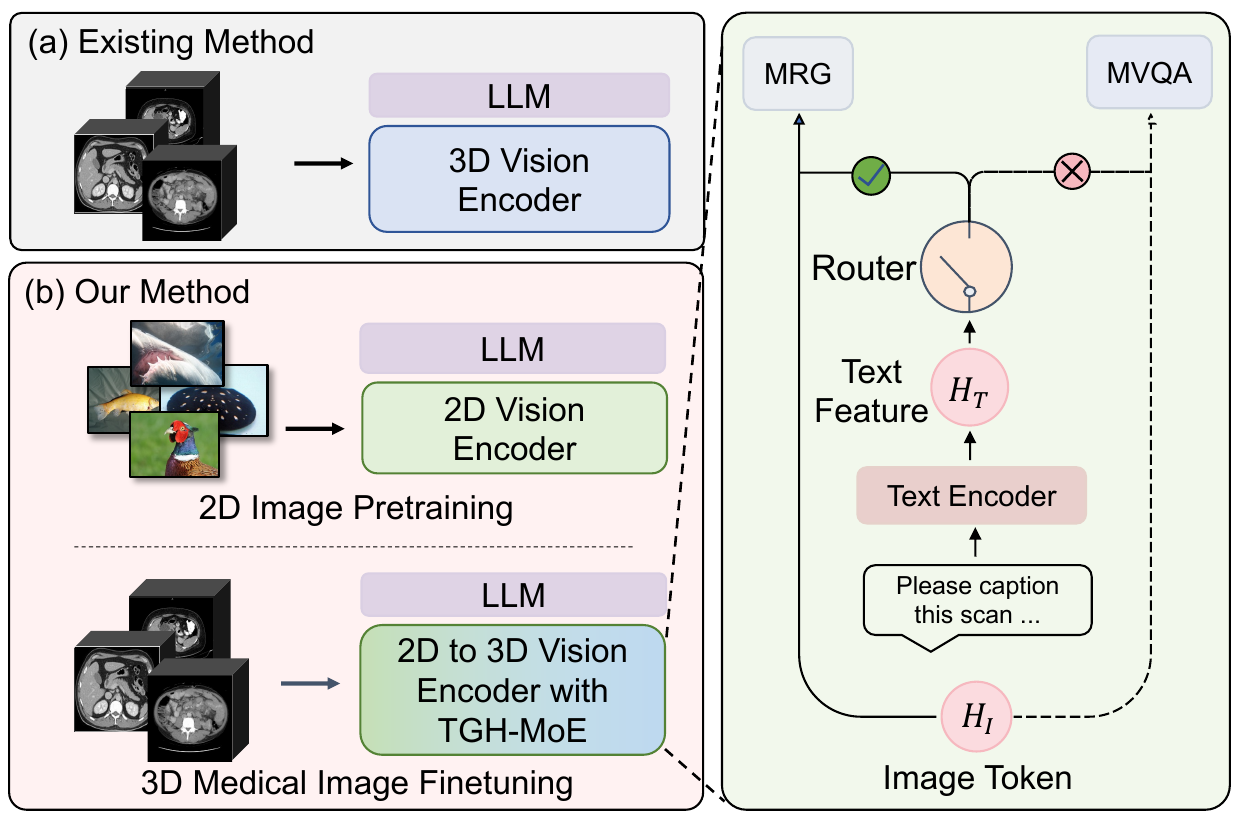}}
\caption{(a) Existing method pre-trains a 3D vision encoder with only 3D medical images in CLIP fashion. It also extracts the same image features for different tasks. (b) Our method adapts a 2D vision encoder to 3D medical images, which is pretrained with massive natural images and aligned with LLMs through instruction-tuning. We further propose TGH-MoE to extract task-specific image features for MRG and MVQA.}
\label{abs}
\end{figure}

In natural image scenarios, numerous 2D vision encoders \cite{zhai2023sigmoid, oquab2023dinov2} pretrained on large-scale image–text corpora exhibit remarkable feature extraction abilities.
2D MLLMs composed of these 2D vision encoders and large language models (LLMs) have shown great capacities in 2D image understanding and instruction following. 
Therefore, a stream of medical MLLMs \cite{yang2025spatio,wang2024surgical} perform tasks such as MRG and MVQA by finetuning these 2D MLLMs with medical image-text pairs. 
For example,  Flamingo-CXR \cite{tanno2025collaboration} adapts the Flamingo vision-language foundation model via fine-tuning, enabling it to produce radiology reports based on chest X-ray (CXR) images.
Nevertheless, such 2D medical MLLMs are unable to effectively extract  3-dimensional spatial information within 3D medical images, such as MRI and CT. To adapt 2D MLLMs for 3D medical images, some works \cite{wang2024interpretable, li2023llava} choose to select key slices from 3D medical images for 2D medical MLLMs to process, sacrificing volumetric context. 
To better utilize 3D spatial information, a few preliminary works \cite{bai2024m3d, wu2025towards} propose to use a 3D ViT as the vision encoder to extract image features. However, unlike the 2D ViT, the 3D ViT can not be pretrained on massive natural 2D image-text pairs, which do not contain 3D spatial information. To enhance the perceptual capacity of the 3D ViT, they propose to collect datasets of 3D medical images and text for pre-training. 
For example,
M3D\cite{bai2024m3d} collects massive CT images along with their corresponding reports as text data, and pre-trains a 3D ViT from scratch by a CLIP-style \cite{radford2021learning} contrastive objective. However, we find that the vision encoders of current 3D medical MLLMs are insufficient for extracting adequate image features to handle MRG and MVQA tasks. 
 In this paper, we identify two progressive levels of limitation that contribute to this insufficiency, regarding fundamental perceptual capacity and advanced multifunctional capacity of the vision encoder. 
For the first level, the perceptual capacity of current MLLMs with 3D ViTs, which are pre-trained solely using 3D medical image-text data, is limited compared to MLLMs with 2D ViTs pre-trained on natural image-text data. This disparity arises because the available medical image-text data is significantly smaller than the natural image-text data by orders of magnitude (tens of thousands vs. billions). 
For the second level, the vision encoders of existing 3D medical MLLMs do not consider the variability in image features required for different tasks, such as MRG and MVQA. Consequently, they extract the same image features for both tasks. In general, MRG requires perceiving the entire image to generate a descriptive report, while MVQA often focuses on a few specific image regions relevant to the question. Failing to consider this difference may lead to suboptimal performance in these tasks.

To tackle the first limitation, we aim to tailor a well-trained 2D MLLM with 2D ViTs to understand 3D medical images as shown in Fig\ref{abs}(b). A well-trained 2D MLLM shows strong perceptual capacity and good image-text alignment. However, how to effectively adapt well-trained 2D MLLMs with 2D ViTs to capture 3D spatial information remains underexplored. We design a series of strategies to adapt a 2D ViT to 3D volumetric inputs while reusing pre-trained parameters to inherit the 2D ViT's strong perceptual capacity. For 2D-to-3D adaptation, we first reuse pretrained 2D patch embedding layer and 2D attention blocks in lower layers to preserve textural details, and then modify 2D attention blocks into 3D ones to extract 3D volumetric information.

To address the second limitation, learning both task-shared and task-specific image features is essential for the vision encoder.
Mixture-of-Experts (MoEs) technique \cite{jacobs1991adaptive} divides the network into several separate sub-networks, known as experts, with each expert specializing in different aspects of the data. 
Inspired by MoEs technique, we design TGH-MoE and integrate it into the vision encoder. Unlike standard MoEs, our TGH-MoE introduces the textual embeddings of prompt into the extraction of image features. Since the task type, i.e., MRG or MVQA can be inferred from prompt, the router routes image features to task-specific experts by taking the text feature of prompt as input. Specifically, our TGH-MoE consists of two levels of MoE, i.e. task- and token-level MoE. Task-level MoE distinguishes MRG and MVQA and utilizes separate experts to extract task-specific image features.
Token-level MoE serves as an expert within task-level MoE to provide a sparse path toward a larger and more powerful model.
However, since image features are routed to different task-specific experts, directly optimizing TGH-MoE hinders the learning of task-shared image features, leading to suboptimal performance. To mitigate this issue, we propose a two-stage training strategy where the first and second stages train the vision encoder to learn task-shared and -specific image features, respectively.

Our main contributions can be summarized as follows: 

\begin{itemize}
\item We propose a 3D medical MLLM which adapts the powerful multimodal capabilities of a well-trained 2D MLLM for processing 3D medical images.

\item We generalize the 2D ViT of a 2D MLLM to 3D medical images, thereby the strong perceptual capacity of the 2D ViT can be inherited, and integrated into the extraction of 3D volumetric information.

\item We introduce text-guided hierarchical MoE and a corresponding two-stage training strategy to train the vision encoder to extract task-shared and task-specific image features for both MRG and MVQA tasks.

\item 
We conduct extensive experiments on a large 3D medical multimodal dataset to validate the superiority of our method over other state-of-the-art methods, and the effectiveness of each component of our method.

\end{itemize}

\section{Related Works}

\subsection{Medical Large Vision-Language Models}

Large Vision-Language Models \cite{zhou2022learning, zhang2022contrastive} have shown good alignment between vision and language modalities to enhance interaction between them.
Inspired by CLIP \cite{radford2021learning}, a series of works attempt to improve the visual representation of the medical images by aligning the medical visual concept with the medical language concept via contrastive learning. For example,  ConVIRT \cite{zhang2022contrastive} utilizes contrastive learning to learn medical visual representations by exploiting medical images and corresponding descriptive texts. SurVLP \cite{yuan2024procedure} proposes to provide the surgical video encoder with comprehensive language supervision by refining and enriching surgical concepts within descriptive texts. Merlin \cite{blankemeier2026merlin} introduces a 3D VLM for abdomen CT using unstructured radiology reports to supervise the 3D vision encoder. Triad framework~\cite{wang2025triad} employs a similar cross-modal strategy to 3D MRI by utilizing textual metadata about scan protocols as supervisory signals to shape the learned visual feature space. Prima~\cite{lyu2026learning} tackles the structural complexity of neuroimaging by aligning large-scale institutionally archived images with summarized reports, thereby achieving highly generalizable representations for diverse diagnostic tasks. These works align vision and language modalities, but they do not mix visual tokens and textual tokens for better interaction.

LLMs \cite{touvron2023llamaopenefficientfoundation, glm2024chatglmfamilylargelanguage, qwen2025qwen25technicalreport} have shown great capability in text understanding and thus been utilized for medical question answering \cite{mcduff2025towards, singhal2025toward,ma2025medla}.
With visual representation aligned with text representation, another series of works proposes to utilize LLMs to process visual tokens and textual tokens simultaneously to enhance the performance of multi-modal tasks. Specifically, MLLMs first extract visual tokens with the vision encoder and then align visual tokens and textual tokens with several projection layers. Finally, visual tokens concatenated with textual tokens are input into LLMs for generating new textual tokens. BiomedGPT \cite{zhang2024generalist} provides a lightweight generalist vision–language foundation model that is pre-trained using medical images in several image modalities, such as X-ray, pathology images, ultrasound, and CT. Some MLLMs have focused on 3D medical images. M3D \cite{bai2024m3d} develops a 3D MLLM trained with the procedure of CLIP pretraining, projector pretraining, and instruction finetuning. Built on top of M3D, Med3DVLM \cite{xin2025med3dvlm} introduces a vision encoder using decomposed 3D convolutions and uses pairwise sigmoid loss for CLIP pretraining. Med3DVLM also uses two separate projectors to align both low- and high-level visual tokens with textual tokens. Existing 3D MLLMs \cite{bai2024m3d, wu2025towards} pre-train 3D ViTs from scratch, which cannot utilize rich information within massive natural image and text data.
In contrast, we propose to modify a well-trained 2D MLLM to adapt to 3D medical images.

\subsection{Medical Visual Text Generation}
In practical medical diagnosis, it's a great burden for radiologists to first analyze the medical image and then write a report or answer specific questions. To help reduce such a burden, medical MLLMs \cite{li2023dynamic, liu2023medical, rao2025multimodal, li2025towards} have been utilized to generate accurate and relevant text responses based on prompts concerning medical images. Medical report generation (MRG) and medical visual question answering (MVQA) are two important tasks in medical visual text generation. 
The MRG works \cite{johnson2019mimic, chen2020generating, jin2024promptmrg} aim to produce relatively lengthy text that provides a complete description of a medical image, whereas the MVQA works \cite{ye2024gmai, hu2024omnimedvqa, du2025lmt++} target at generating more concise text that directly addresses a specific question about the image. 
 MRG requires the vision encoder to focus on the global features of images. For example, RGRG \cite{tanida2023interactive} first detects several anatomical regions and then generates a description for all regions to get the final report. STREAM \cite{yang2025spatio} generates reports for chest X-ray images by utilizing MLLM to capture spatio-temporal visual dynamics and retrieves semantic entities from a pre-constructed knowledge bank. The generated reports include description of all important organs like lungs, pleura, heart, and so on.
 In contrast, MVQA requires the vision encoder to focus on specific image features. For example, SurVLM \cite{zeng2025surgvlm} constructs a large-scale surgical MVQA dataset by turning multiple datasets of various tasks like instrument recognition into MVQA format. The instrument recognition requires the model to focus on the regions of instruments.  

Some works attempt to develop a model for both MRG and MVQA. RadFM \cite{wu2025towards} integrates several MRG and MVQA datasets into a large medical image dataset containing both 2D and 3D images, and trains a generative foundation model for MRG and MVQA.
Focusing on 3D medical images, M3D \cite{bai2024m3d} presents a large 3D medical image dataset containing 3D images paired with medical reports and utilizes LLMs to create question-answer pairs based on medical reports. M3D pre-trains the 3D ViT of an MLLM by aligning the images embeddings and report embeddings. Then it instruction-tunes the MLLM with both the reports and question-answer pairs. 
Existing works do not consider the task variety between MRG and MVQA, which require different image features. In contrast, we suggest using a text-guided MoE strategy to learn task-shared and -specific image features for MRG and MVQA.

\section{Method}
In this section, we will first present some preliminaries about MLLMs and Mixture-of-Experts. Then we will describe our strategy for adapting a well-trained 2D MLLM to 3D medical images. Next, we will elaborate on our text-guided hierarchical MoE (TGH-MoE) framework. Lastly, we will introduce our two-stage training strategy.

\subsection{Preliminaries}
\subsubsection{Multimodal Large Language Model}
We utilize a well-trained 2D MLLM, which is composed of a 2D ViT as a vision encoder $e_{I}$, a connector $c_{I}$ and an LLM $f_{LLM}$. The input of the MLLM consists of a 3D medical image $X_{I} \in \mathbb{R}^{D \times H \times W}$ and a text prompt $X_{T}$. The vision encoder extracts image features, and the connector aligns the image features with text features:
\begin{equation}
\label{de_kl}
T_I
 = c_I(e_I(X_I)).
\end{equation}
Note that since $e_I$ is originally a 2D encoder, we modify and apply it to the 3D volume $X_I$ in a slice-wise manner, as detailed in Section.\ref{2d-to-3d}. 
The text prompt is converted into text embeddings $T_T$ by a tokenizer and then concatenated with image embeddings $T_I$. The LLM takes $[T_T, T_I]$ as input and outputs a textual response $Y_R$ to address the question within the text prompt.

\subsubsection{Mixture-of-Experts}
The Mixture-of-Experts (MoE) model \cite{shazeer2017outrageously} proposes to utilize a collection of specialized expert networks $\epsilon = [e_1, e_2, ..., e_M]$, each designed to capture distinct aspects of information.  
Additionally, the model incorporates a router that dynamically assigns different subsets of these expert networks to specific tasks based on their unique requirements. The router takes an indicator vector $v$ as input and outputs the weights $w \in \mathbb{R}^{M}$ corresponding to these experts:
\begin{equation}
\label{router}
w
 = Softmax(W_T\cdot v).
\end{equation}
Here, $W_T \in \mathbb{R}^{M \times d_T}$ is a learnable transformation matrix. Finally, the MoE function can be formulated as the combination of experts:
\begin{equation}
\label{moe}
MoE(h, v)
 = \sum_{i=1}^{M} w_ie_i(h).
\end{equation}

\subsection{Overall Framework}
In this work, we propose a novel 3D MLLM that focuses on the design of its vision encoder derived from the 2D ViT of a well-trained 2D MLLM. However, the 2D ViT cannot directly learn 3D spatial information from 3D medical images. Also, it cannot extract specific image features for different medical tasks. To address these limitations, we first thoroughly analyze all components of a 2D ViT and carefully tailor these components for 3D medical images. Then we propose a text-guided hierarchical MoE to extract task-specific image features, as shown in Fig.\ref{fig2} (c). Finally, we devise a two-stage training strategy which enables TGH-MoE to learn both task-shared and -specific image features.  

\subsection{Adapting 2D MLLM for 3D Medical Images} \label{2d-to-3d}
Although a 3D vision encoder can seamlessly process 3D volumetric inputs, its perceptual capacity is limited by insufficient pre-training caused by the scarcity of 3D medical images. Meanwhile, it has not been connected to an LLM to go through instruction tuning with natural image and text data, leading to a poor alignment with the LLM. 
In contrast, a well-trained 2D MLLM exhibits more powerful perceptual capacity and strong alignment between image and language information. But it cannot understand 3D volumetric information. To address these limitations, we propose to adapt a well-trained 2D MLLMs to process 3D medical images. 
\begin{figure}[!t]
\centerline{\includegraphics[width=\columnwidth]{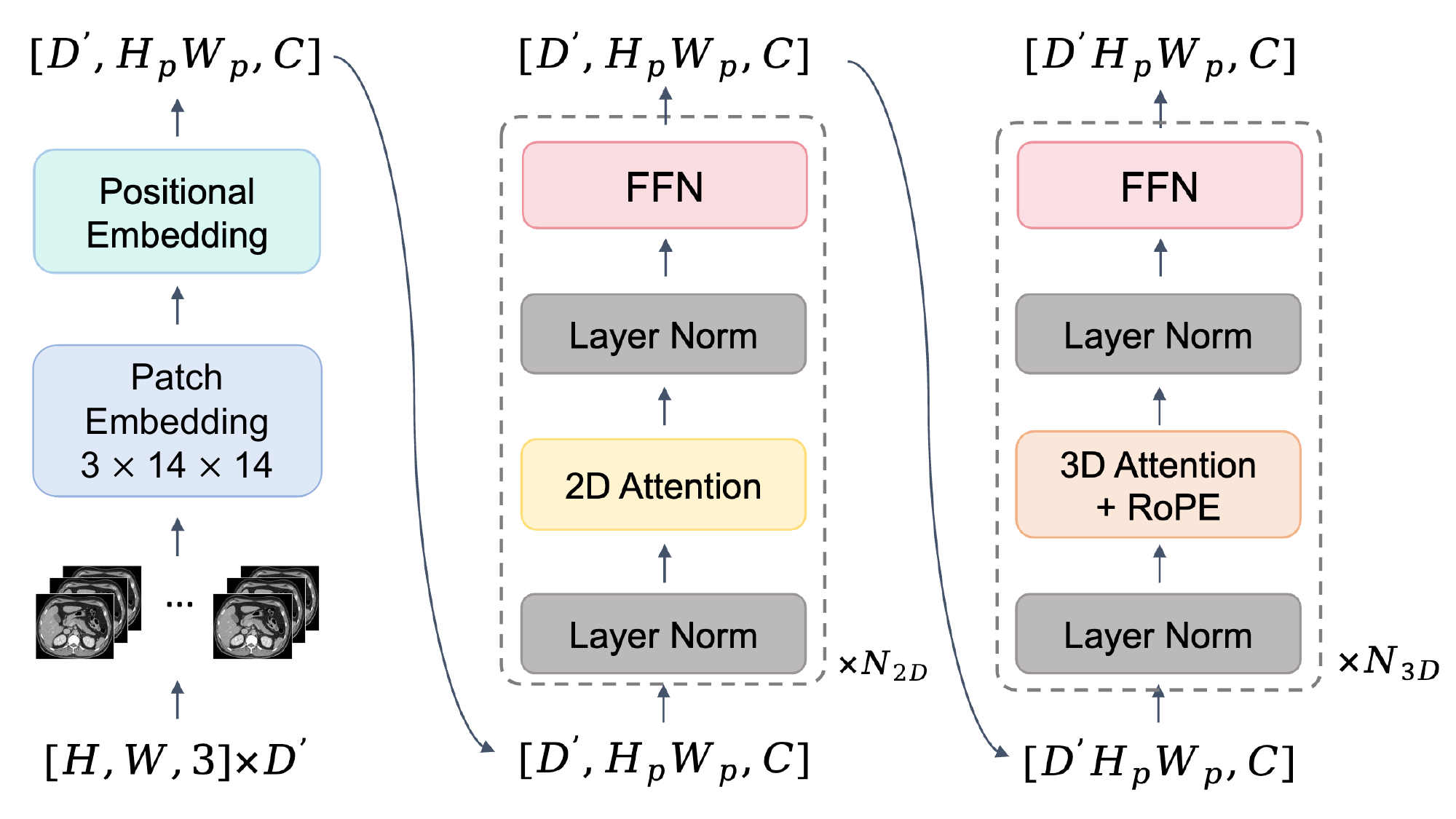}}
\caption{Illustration of adapting 2D ViT for 3D medical images. We split a 3D image into multiple 3-slice images for patch embedding processing. We modify 2D attention blocks in higher layers into 3D attention blocks to capture 3D spatial information. RoPE is added to 3D attention blocks to distinguish positions along the $D$ dimension.}
\label{fig1}
\end{figure}

Most open-sourced 2D MLLMs take a 2D ViT as the vision encoder. We first thoroughly analyze all components of the 2D ViT and then customize these components for 3D medical images. Within the 2D ViT, an image is first converted to image tokens by the patch embedding layer. Then, positional embedding is added to the image tokens to mark their position in the image. Next, attention blocks update image tokens by learning relations between these tokens. After certain attention blocks, patch merging layers are deployed to reduce image tokens. As shown in Fig.\ref{fig1}, we identify key gaps between these components and 3D medical images and accordingly modify them to adapt to 3D medical images:

\subsubsection{Patch Embedding} The original 2D ViT takes a 2D convolution layer with 3 channels as the patch embedding layer. As a result, it cannot process 3D medical images having more than 3 slices.
To adapt the patch embedding layer for 3D medical images, we treat a 3D volume as a sequence of 2D slices. Specifically,  we resize a 3D medical image of shape $[D, H, W]$ into $[D^{\prime}, 3, H, W]$, where $[D = D^{\prime} \times 3]$. Then the patch embedding layer can directly process it into embeddings of shape $[C, D^{\prime}, H_p, W_p]$. 

\subsubsection{Positional Embedding} The original 2D ViT can only distinguish positions along $H$ and $W$ dimensions by using a group of pretrained parameters of shape $[C, H_p, W_p]$ as 2D positional embeddings. However, positions along the $D$ dimension cannot be distinguished if only 2D positional embeddings are added. To address this limitation, rotary position embedding (RoPE)\cite{su2024roformer}, which contains no learnable parameters, is added to queries and keys in self-attention layers. We also repeat and add the original 2D positional embeddings to patch embeddings.  

\subsubsection{Attention Block} The original 2D ViT contains only 2D attention blocks which only extract image features along H and W dimensions, and ignore information along D dimension. We extend 2D attention to 3D attention by reshaping the input from $[D^{\prime}, H_pW_p, C]$ to $[D^{\prime}H_pW_p, C]$ and make no modification on pretrained parameters within attention blocks. 

\subsubsection{Patch Merging} 
Since the original patch merging layers merge four adjacent tokens within a 2x2 block on a 2D plane into one token by using linear layers with learnable parameters. We reuse these patch merging layers along the $H$ and $W$ dimensions. To further reduce the number of image tokens, we apply an average pooling operation along the D dimension with a kernel size of 2 after the last 2D patch merging layer.

With the above mentioned modifications, the 2D ViT is able to process 3D medical images. 
An existing study \cite{raghu2021vision} indicates that for a well-trained ViT, its lower layers tend to focus more on local information, while higher layers primarily incorporate global information. 
Since the 2D ViT is pre-trained with massive 2D images, its lower layers are specialized in capturing 2D local information. 
In contrast, higher layers with the capacity to incorporate global information are more suitable to extract 3D volumetric information. Therefore, we propose to keep lower layers unchanged and modify 2D attention blocks in higher layers into 3D attention blocks, which incurs a mixture of 2D and 3D attention blocks, as shown in Fig.\ref{fig1}. Compared with the vision encoder with full 3D attention blocks, our strategy 
can achieve slightly higher performance even with less computational cost. Since the original 2D ViT is not pre-trained with 3D medical images, it struggles with capturing 3D volumetric information without fine-tuning. Therefore, we propose to keep its parameters updated during finetuning the MLLM, instead of keeping it frozen like in existing methods \cite{bai2024m3d, wu2025towards}.

Our adaptation exhibits four advantages: (1) enabling a given 2D ViT to process 3D volume images effectively; (2) fully leveraging the pre-trained model by reusing all pre-trained parameters; (3) introducing no new learnable parameters to reduce training difficulty; (4) taking less computational cost than full 3D attention with better performance.

\begin{figure*}[ht]
\centering
    \includegraphics[width=\textwidth]{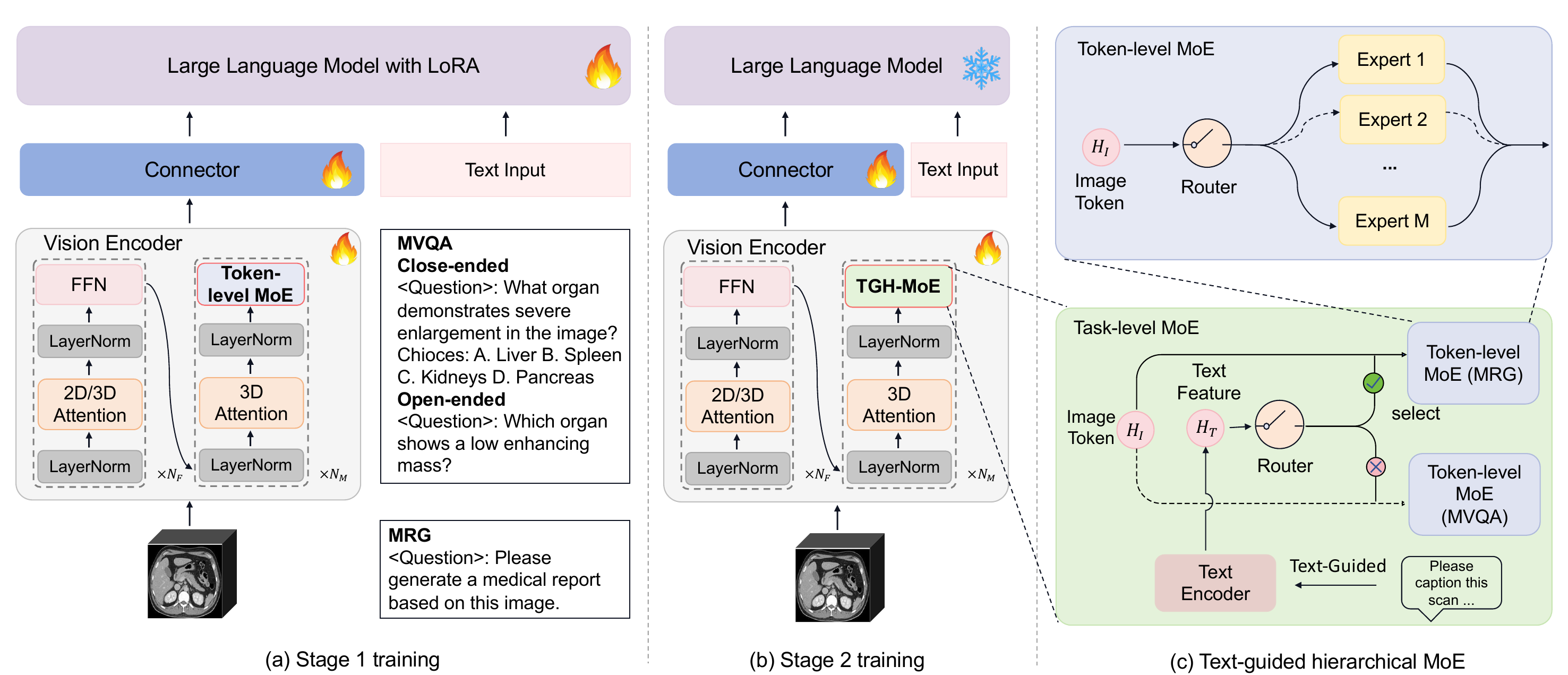}
    \caption{ We propose a text-guided hierarchical MoE (TGH-MoE) and a two-stage training strategy to learn task-shared and -specific image features for MRG and MVQA tasks. (a) In Stage 1, we use only token-level MoE so that the vision encoder can learn task-shared features for MRG and MVQA.  (b) In Stage 2, we replicate token-level MoE as the experts of task-level MoE to learn task-specific image features for MRG and MVQA under the guidance of text prompt. 
    (c) TGH-MoE is a two-level MoE. Task-level MoE first routes all image tokens to a specific expert under the guidance of the text prompt. Token-level MoE provides a sparse path toward a larger model by routing image tokens to different experts. }
\label{fig2}
\end{figure*}

\subsection{Text-guided Hierarchical MoE}
It's important for a medical MLLM to achieve robust performance for both MRG and MVQA tasks. MRG requires the vision encoder to perceive the overall image, while MVQA focuses on specific regions relevant to a given question. However, existing works extract the same image features for various tasks, which cannot distinguish the difference between the two tasks regarding image features.
We argue that task type i.e., MVQA or MRG can be deduced from the text prompt $X_T$. Therefore, we take such text information into consideration when extracting image features for different tasks. 
 
 Based on MoEs technique, which repeats the sub-part of a network for various tasks, we design a text-guided hierarchical MoE (TGH-MoE) framework to extract task-specific image features. As shown in Fig.\ref{fig2}(c), our text-guided hierarchical MoE employs a two-level routing mechanism, comprising a task-level MoE and a token-level MoE. The task-level MoE first routes the entire set of image tokens based on the text prompt so that task-specific image features are learned. Subsequently, within each of these task-specific experts, we use a token-level MoE to create a sparser and stronger network. The token-level MoE can better capture image features for each individual image token by learning a specialized combination of features.

The task-level MoE takes the text prompt $X_T$ as the input of the router to guide the selection of experts, because task type i.e., MVQA or MRG can be deduced from $X_T$. Specifically, we employ the first four layers of our LLM as text encoder $e_T$ to encode $X_T$ into an indication vector and derive the weights $wt$ of experts following Eq.\ref{router}:
\begin{equation}
\label{ht}
wt
 = Softmax(W_T \cdot e_T(X_T)).
\end{equation}
A router can be categorized into a soft and hard one based on whether we binarize the weights $wt$. A soft router does not binarize $wt$, and combines the outputs of all experts. In contrast, recent works \cite{li2023pace, riquelme2021scaling} have proposed hard routers, which pre-assign each expert to a specific task. A key advantage of the hard router is its mitigation of optimization interference during multi-task training. To learn task-specific image features, we assign two experts for MVQA and MRG, respectively, and employ a hard router to select between them by binarizing the weights $w_t$ based on:
\begin{equation}
\label{binary}
wt^\prime_i = \left\{
\begin{aligned}
&1, & argmax(wt) = i \\
&0, & argmax(wt) \neq i.\\
\end{aligned}
\right.
\end{equation}
Once a task-specific expert is selected, all image tokens will be processed by this expert. We enable the hard router to classify the text prompts into MVQA or MRG by training it with cross-entropy loss $H$ :
\begin{equation}
\label{router_loss}
L_{r}
 = H(yt, wt).
\end{equation}
Note that we stop the gradient of the indication vector $h_T$ so that $L_{r}$ will not interfere with the training of the text encoder $e_T$, which is a part of LLM.

Within the task-level MoE, each task-specific expert is set to be a token-level MoE to create a sparse path toward a larger and more powerful model. To instantiate multiple token-level experts, we replicate FFN for $M$ times. For the token-level MoE, the expert weights are computed by feeding the image token $h_I$ into a soft router. This router selects the top-K experts and applies a softmax function for normalization:
\begin{equation}
\label{router2}
wk
 = Softmax(Top(W_T\cdot h_T, K)).
\end{equation}

\begin{equation}
\label{router3}
Top(x, K) = \left\{
\begin{aligned}
&x_i, & if \ x_i\ \in topK(x)\\
&0, & otherwise.\\
\end{aligned}
\right.
\end{equation}
Each image token corresponds to a group of weights, resulting in a specific combination of experts according to Eq.\ref{moe}.

As shown in Fig.\ref{fig1}, our vision encoder consists of stacked attention blocks, and each attention block has a feed-forward neural network (FFN). Inspired by \cite{gou2023mixture, jiang2024med, lin2024moe}, we replace the FFNs in certain attention blocks with our TGH-MoE.
To balance shared representation and task specialization, only the FFNs in the last six attention blocks are replaced. This design allows MVQA and MRG tasks to share low-level image features while enabling them to diverge in their high-level feature representations.

\subsection{Training Strategy}
As shown in Tab.\ref{tab:component}, directly optimizing our TGH-MoE alongside the fine-tuning of the entire MLLM leads to suboptimal performance. We attribute this to the fact that direct optimizing is insufficient to capture task-shared image features. This occurs because the task-level MoE routes features to task-specific experts from the beginning of training, preventing the learning of a common representation. To enable our TGH-MoE to learn both task-shared and -specific image features, we propose a two-stage training strategy. In Stage 1, our objective is to train the MoE module to learn task-shared features. As shown in Fig.\ref{fig2}(a), in Stage 1, we use only the token-level MoE and abandon the task-level MoE, forcing both MRG and MVQA tasks to share the same image features. We treat all MRG and MVQA data as instruction data to train the whole network using an auto-regressive loss:
\begin{equation}
\label{re_loss}
L_{reg}
 = H(y^i, p_\theta(\hat{y}^{i}|V, \hat{y}^{[:i-1]})).
\end{equation}
During Stage 1, as illustrated in Fig. \ref{fig2}(a), the vision encoder, connector, and LLM are all optimized to adapt to the medical multimodal data. We apply the LoRA strategy \cite{hu2022lora} to the LLM to make our training more efficient.

In Stage 2, our objective is to train the TGH-MoE to learn task-specific image features. We initialize the two experts for the task-level MoE by replicating the token-level MoE trained in Stage 1. We again use all MRG and MVQA data for instruction tuning with the auto-regressive loss.
In contrast to Stage 1, we keep the LLM frozen and focus on fine-tuning only the vision encoder and the connector to learn task-specific features.
As shown in Fig.\ref{fig2} (c), the router of task-level MoE will route all image tokens of an image to a specific expert according to its text prompt. We train the router with $L_r$ in Eq.\ref{router_loss} to distinguish the two tasks. The loss in Stage 2 can be formulated as:
\begin{equation}
\label{total_loss}
L_{total}
 = L_{reg} + \alpha L_r,
\end{equation}
where $\alpha$ is a hyperparameter to control the weight of $L_r$.

\begin{table*}[ht]
\begin{center}
\caption{Comparison with existing methods. All results are reported in percentages ($\%$). For Medical VQA, we report the average of five different topics. $\dagger$ indicates that the existing methods are not trained on M3D-Data, while $\star$ is on the contrary.}
\begin{tabular}{c|c|ccccc|cccc}
\hline
\multirow{2}{*}{Method} & \multirow{2}{*}{Model Size}  & \multicolumn{5}{c|}{Medical VQA}   & \multicolumn{4}{c}{Report Generation}       \\
                                                  &  & BLEU-1   & ROUGE-1   & METEOR     & BERT  & Accuracy  & BLEU-1      & ROUGE-1     & METEOR       & BERT     \\ \hline
                    LLAVA-Med-1.5 $\dagger$  &  7B       & 34.53 &36.54  & 25.43  & 89.06  & 64.41                         & \textbf{14.25}    & 16.08  & 12.74   & 82.53       \\
                    
                    RadFM $\dagger$  &  14B      & 16.39 & 26.13 & 21.33 & 88.72 & 19.79                       & 12.23  & 16.49   & 11.57  & \textbf{87.93}       \\
M3D $\star$   &  4B               & 52.10 & 55.19 & 35.89 & 92.14 & 79.35        & 13.90  & 18.02   & 12.77  & 83.63       \\
Med3DVLM $\star$  & 7B                  & 47.54 & 50.98 & 32.68 & 91.32 & 76.89         & 13.86       & 18.01        & 12.64       & 83.70       \\
Ours  & 4B                & \textbf{54.64} & \textbf{57.97} & \textbf{37.85} & \textbf{92.57} & \textbf{81.06}            & 14.23       & \textbf{19.03}        & \textbf{13.40}       & 84.03      \\ \hline
\end{tabular}
\label{tab:main}
\end{center}
\end{table*}

\section{Experiments}

\subsection{Datasets}
To ensure the generalizability of our approach, we utilize both caption and Visual Question Answering (VQA) data from M3D-Data \cite{bai2024m3d}
, a large-scale, multimodal 3D medical dataset \footnote[1]{Our usage is approved by Radiopaedia with ID 202407-0012.} encompassing a diverse range of organs. The caption data comprises concise reports that provide overall descriptions of the images. From these reports, VQA samples are automatically generated by LLMs, with each question focusing on a specific medical aspect of the image. The VQA data is categorized into five distinct topics: plane, phase, organ, abnormality, and location, enabling a targeted evaluation of the model's performance across various facets of medical image understanding. The VQA questions are presented in two formats: open-ended and closed-ended. Open-ended questions require a free-form answer, whereas closed-ended questions provide multiple choices for the model to select from.

The M3D-Data dataset \cite{bai2024m3d} provides over 120k caption samples and 450k VQA samples for training, along with 2k caption samples and 27k VQA samples for testing. However, we identified 1,666 test caption samples that overlap with the training set, sharing identical captions. Since the VQA samples are generated from these captions, some test VQA samples could potentially be answered using information from the training captions. To ensure a fair comparison, we filter out all overlapping samples from the training data to create a strictly held-out test set. This results in a final dataset of 106,192 caption samples and 437,953 VQA samples.

To ensure a fair comparison, we employ the same test set as M3D \cite{bai2024m3d}. Following M3D's evaluation protocol, we use accuracy to assess performance on the closed-ended VQA task.
For the report generation (MRG) and open-ended VQA tasks, we evaluate the quality of the generated text using standard natural language generation (NLG) metrics. These include BLEU-1 (B-1), ROUGE-1 (R-1), METEOR, and BERT-Score, which measure the similarity between the generated text and the ground-truth reference.

\subsection{Implementation Details}
We use Phi-3-V \cite{abdin2024phi3technicalreporthighly} as our MLLM, aligning with the SOTA method M3D \cite{bai2024m3d} which uses the same LLM. Phi3-V takes a 24-layer 2D ViT as vision encoder. The input image resolution is $24 \times 336 \times 336$. The number of attention layers is configured as $N_{2D}=6$ for 2D and $N_{3D}=17$ for 3D. We deploy TGH-MoE in the last 4 layers of the vision encoder, leading to $N_M=4$ layers with TGH-MoE and $N_F=19$ layers without TGH-MoE. For token-level MoE, we set the total number of experts $M$ to 4 and activate the top 2 experts for each token. We implement our method and benchmark baselines in PyTorch. Our model is trained using the AdamW optimizer with learning rate of $2e-5$ and cosine learning rate scheduler. We enable BF16 mixed-precision training via DeepSpeed \cite{rajbhandari2020zero} to enhance efficiency. We conduct our experiment using 8 NVIDIA A40 GPUs, each with 46 GB of memory.

\subsection{Comparison with Other Methods}
\subsubsection{Evaluation on Medical Visual Question Answering}
As shown in Tab.\ref{tab:main}, we compare our methods with LLAVA-Med-1.5 \cite{li2023llava}, RadFM \cite{wu2025towards}, M3D \cite{bai2024m3d} and Med3DVLM \cite{xin2025med3dvlm}. We evaluate LLAVA-Med-1.5 and RadFM using their publicly available pre-trained weights from HuggingFace. For M3D and Med3DVLM, we train the models on our processed training data using their official code bases.
Since LLAVA-Med-1.5 is designed for 2D images, we preprocess 3D medical volumes by extracting 9 equally spaced slices and arranging them into a 3×3 grid for input. For the VQA task, we report the average performance across the five distinct topics.

Our proposed method outperforms existing approaches by a significant margin on both open- and closed-ended VQA tasks. On the closed-ended task, our method achieves an accuracy of $81.06\%$, surpassing the second-best method, M3D ($79.35\%$), by $1.71\%$. For open-ended VQA, our method also exceeds M3D's performance, improving BLEU and ROUGE scores by $2.54\%$ and $2.78\%$, respectively.
These superior results demonstrate that our method possesses a stronger capability to interpret 3D medical images and generate more accurate answers to questions about specific aspects of an image.

\subsubsection{Evaluation on Medical Report Generation}
Our method achieves the best results in ROUGE and METEOR scores, indicating its superior performance in generating text with higher textual quality and clinical completeness. While LLAVA-Med-1.5 achieves a slightly higher BLEU-1 score than our method, it performs significantly worse across all other metrics. However, BLEU-1 only reflects the precision of the prediction and it cannot reflect the recall like ROUGE-1. With much higher recall and slightly lower precision, our methods outperform LLAVA-Med-1.5.
Although RadFM attains a higher BERT score, our method substantially outperforms it in the remaining eight out of nine metrics. Crucially, our model delivers better overall performance than RadFM while using only 4B parameters, compared to its 14B.

Our method's superior performance on both MRG and MVQA underscores its overall effectiveness. Both M3D and Med3DVLM employ 3D vision encoders pre-trained on 3D medical images, and their models come with a similar or even larger parameter count. However, their image features prove insufficient for the distinct demands of MRG and MVQA. 
In contrast, our method adapts the vision encoder from a pre-trained 2D MLLM for 3D medical images, thereby leveraging its strong perceptual capacity and superior image-text alignment. Furthermore, we propose a Text-Guided Hierarchical Mixture of Experts (TGH-MoE) framework, which enables the vision encoder to learn both task-shared and task-specific image features, effectively addressing the challenges posed by task variety. As a result, our approach successfully extracts the high-quality image features necessary to excel at both tasks.

\begin{table*}[ht]
\begin{center}
\caption{Components analysis of our method. We conduct this ablation to illustrate the importance of our three key components, i.e., 2D to 3D adaptation, TGH-MoE and two-stage training strategy.}
\begin{tabular}{cccc|ccccc|cccc}
\hline
\multirow{2}{*}{No.} & \multirow{2}{*}{2D to 3D Adap.} & \multirow{2}{*}{TGH-MoE} & \multirow{2}{*}{Two-stage Train. Stra.} &   \multicolumn{5}{c|}{Medical VQA} &\multicolumn{4}{c}{Report Generation}\\
                    &      &                          &             & B-1   & R-1   & M     & BERT  & Accuracy                       & B-1      & R-1     & M       & BERT     \\ \hline
                    1&      &                          &                  & 52.10 & 55.35 & 35.62 & 92.15 & 78.41                  & 14.67    & 19.54   & 13.97   & 83.88       \\
                    2&      & $\checkmark$                        &              & 52.83      & 56.05      & 36.44      & 92.30      & 78.96                       & 14.78         & 19.48        & 14.08        &83.94                 \\
                    3&      & $\checkmark$                        & $\checkmark$                & 52.87 & 56.08 & 36.43 & 92.29 & 79.11                  & 14.73    & 19.87   & 13.87   & 83.98       \\
4& $\checkmark$                         &                          &                      & 52.74 & 55.95 & 36.20 & 92.30 & 79.13              & 14.69    & 20.02   & 14.09   & 84.05       \\
5& $\checkmark$                         & $\checkmark$                        &                 & 53.02 & 56.17 & 36.52 & 92.33 & 79.28                   & 14.91    & 19.55   & 14.11   & 83.92       \\
6& $\checkmark$                         & $\checkmark$                        & $\checkmark$             & \textbf{53.10} & \textbf{56.39} & \textbf{36.61} & \textbf{92.33} & \textbf{79.32}                     & \textbf{15.10}     & \textbf{20.31}   & \textbf{14.23}   & \textbf{84.12}       \\ \hline
\end{tabular}
\label{tab:component}
\end{center}
\end{table*}

\subsection{Ablation Study}
To reduce training cost, we use all caption training samples and $1/3$ randomly selected VQA training samples to conduct our ablation studies.

\begin{table*}[ht]
\begin{center}
\caption{Ablation on training-strategy. We compare our training strategy with strategies that can learn only task-specific or shared image features. For ``Only Task-shared", we abandon task-level MoE so that different tasks share the same expert which is a token-level MoE. For ``Only Task-specific", we skip Stage 1 training and keep Stage 2 training.}
\begin{tabular}{ccc|ccccc|cccc}
\hline
\multicolumn{3}{c|}{Design Choices}                            & \multicolumn{5}{c|}{Medical VQA}   & \multicolumn{4}{c}{Report Generation}                                                 \\
Description       & $N_{Stage}$ & MoE Type          & B-1            & R-1            & M     & BERT           & Accuracy                      & B-1   & R-1            & M     & BERT                  \\ \hline
Only Task-shared   &1 &Token-level        & 52.77          & 56.05          & 36.29 & 92.29          & 79.24     & 14.96 & 20.03          & 14.07 & 84.06                    \\
Only Task-specific   &1 &TGH-MoE     & 53.02 & 56.17 & 36.52 & 92.33 & 79.28      & 14.91    & 19.55   & 14.11   & 83.92             \\
Finetune LLM in Stage 2 &2 &TGH-MoE     & 52.73          & 55.98          & 36.47 & 92.26          & 79.29   & 15.10 & 19.62          & 13.91 & 83.89                    \\
Task-shared and -specific (Ours)  &2 &TGH-MoE         & \textbf{53.10} & \textbf{56.39} & \textbf{36.61} & \textbf{92.33} & \textbf{79.32}                 & \textbf{15.10}     & \textbf{20.31}   & \textbf{14.23}   & \textbf{84.12}    \\ \hline
\end{tabular}
\label{tab:training-strategy}
\end{center}
\end{table*}

\subsubsection{Component Analysis.}
We conduct ablation studies to evaluate the effectiveness of our core components: the 2D-to-3D adaptation, the Text-Guided Hierarchical MoE (TGH-MoE), and the two-stage training strategy.
For the experiments labeled ``w/o 2D-to-3D adaptation," we maintain all attention blocks in the vision encoder in their original 2D form. Therefore, the self-attention mechanism only operates across 2D slices, preventing the model from extracting features along the depth $D$ dimension and thus failing to fully exploit the spatial information within 3D medical images. As shown in Tab.\ref{tab:component}, the comparison of the results (e.g., No. 1 vs. No. 4, No. 2 vs. No. 5, and No. 3 vs. No. 6) demonstrates that models with the 2D-to-3D adaptation consistently outperform their counterparts without it across all evaluation metrics for MRG, open-ended MVQA, and closed-ended MVQA. This result confirms the critical importance of effectively utilizing 3D spatial information.

For experiments without ``TGH-MoE", we keep the original feed-forward networks (FFNs) in the attention blocks unchanged, forcing MRG and MVQA tasks to share identical image features. For experiments without ``Two-stage training strategy", we train the model end-to-end in a single stage.
Comparisons between ``No. 1 vs. No. 2" and ``No. 4 vs. No. 5" reveal that employing the TGH-MoE without the two-stage strategy improves performance on open- and closed-ended MVQA but fails to deliver consistent gains for MRG. We attribute this suboptimal outcome to the fact that, without the two-stage strategy, the model is unable to learn task-shared image features, as features are immediately routed to task-specific experts from the start of training.
In contrast, our proposed two-stage strategy first trains the experts to learn task-shared features in Stage 1. These experts are then used to initialize the task-specific experts in Stage 2, which focuses on learning task-discriminative features. As shown by the comparisons ``No. 1 vs. No. 3" and ``No. 4 vs. No. 6," combining the TGH-MoE with the two-stage training strategy leads to substantial performance improvements across all tasks: MRG, open-ended MVQA, and closed-ended MVQA.

\begin{figure}[!t]
\centerline{\includegraphics[width=\columnwidth]{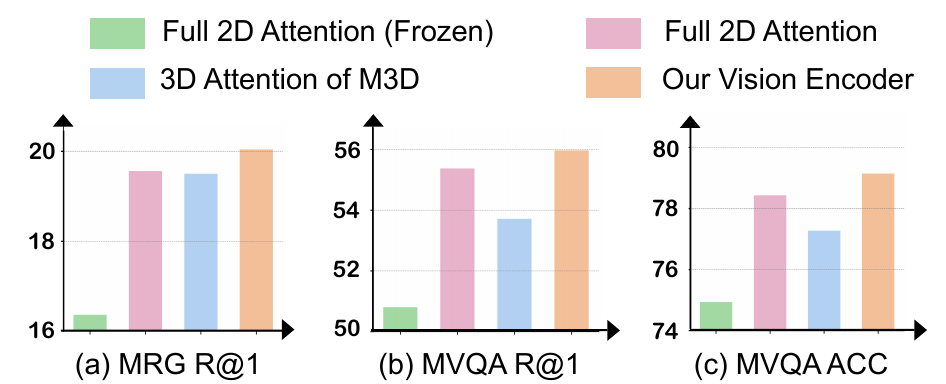}}
\caption{Ablation on the choice of the vision encoder. We compare the frozen and optimized vision encoder during training. We also compare our vision encoder with other 2D and 3D alternatives.}
\label{bar}
\end{figure}

\subsubsection{Ablation on Choice of the Vision Encoder}
The original ViT within the MLLM is pre-trained on 2D natural images, which exhibit a significant domain gap compared to 3D medical images. Therefore, we optimize the ViT alongside the LLM rather than keeping its parameters frozen. As shown in Fig. \ref{bar}, using a frozen vision encoder leads to a substantial performance drop across all metrics, underscoring the importance of fine-tuning the vision encoder for the target domain.

Next, we compare our vision encoder, adapted from a 2D ViT, against the original 2D ViT (Full 2D Attention in Fig. \ref{bar}) and the 3D ViT used in M3D \cite{bai2024m3d}. For the original 2D ViT baseline, we follow the same training recipe as our adapted encoder. To ensure a fair comparison with the 3D ViT, we replace our vision encoder with the 3D ViT while keeping the LLM initialized with the same parameters as in our model. This 3D ViT is pre-trained on caption data using a CLIP-like objective. 
As shown in Fig. \ref{bar}, the original 2D ViT outperforms the 3D ViT, which emphasizes the importance of pre-training with massive 2D images. Compared with the original 2D ViT, our vision encoder achieves higher results,  demonstrating the effectiveness of our adaptation strategy. Fig. \ref{bar} shows that our vision encoder achieves superior performance compared to the 3D ViT. This result indicates that a 2D ViT, once adapted and leveraging its pre-training on massive natural images, possesses stronger perceptual capacity than a 3D ViT pre-trained solely on a limited set of 3D medical images.
 
\begin{figure}[!t]
\centerline{\includegraphics[width=0.85\columnwidth]{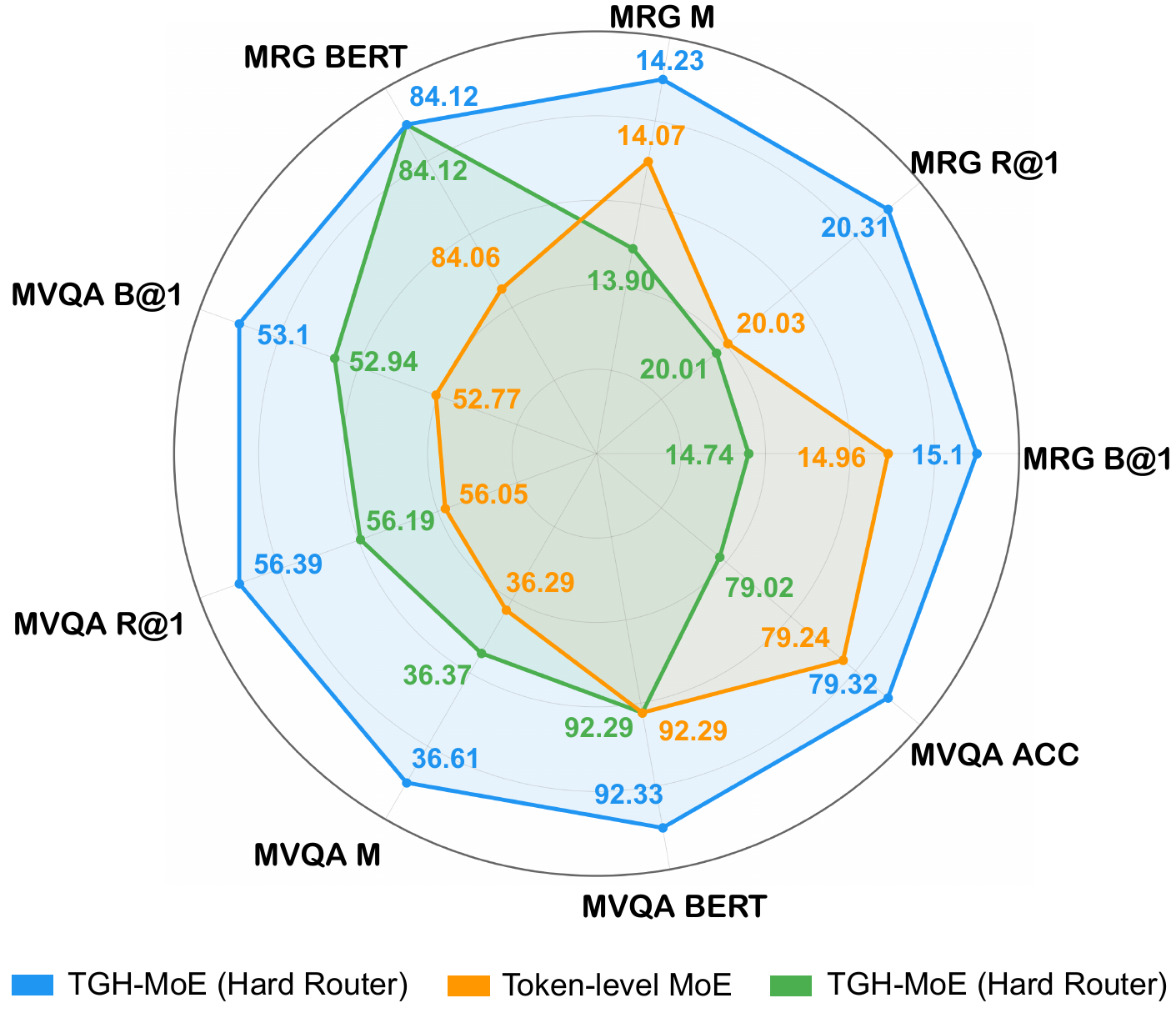}}
\caption{Ablation on TGH-MoE. We compare our TGH-MoE using a hard router with TGH-MoE using a soft router and token-level MoE.}
\label{radar}
\end{figure}

\subsubsection{Ablation on TGH-MoE} 
Our TGH-MoE learns task-specific image features guided by the text prompt. In this ablation study, we compare our TGH-MoE against two alternative MoE architectures.
The first alternative is a ``Token-level MoE," which uses image tokens as the router's input. This design provides a sparse pathway to a larger model but does not account for the fundamental differences between the MRG and MVQA tasks. As shown in Fig. \ref{radar}, our TGH-MoE outperforms the Token-level MoE, demonstrating the benefit of explicitly modeling task-specific features.
The second alternative is a ``TGH-MoE with a soft router," which replaces the hard router in our task-level MoE with a soft router. In this setup, image features are updated according to Eq. 3, where all experts contribute to the feature update for a given task. Consequently, the tasks may interfere with each other, increasing the learning difficulty. As shown in Fig. \ref{radar}, the soft-router variant performs worse than our hard-router TGH-MoE across all metrics and is even inferior to the Token-level MoE on most metrics. These comparisons underscore the importance of using a hard router to isolate tasks and prevent optimization interference.

\begin{figure*}[ht]
\centering
    \includegraphics[width=\textwidth]{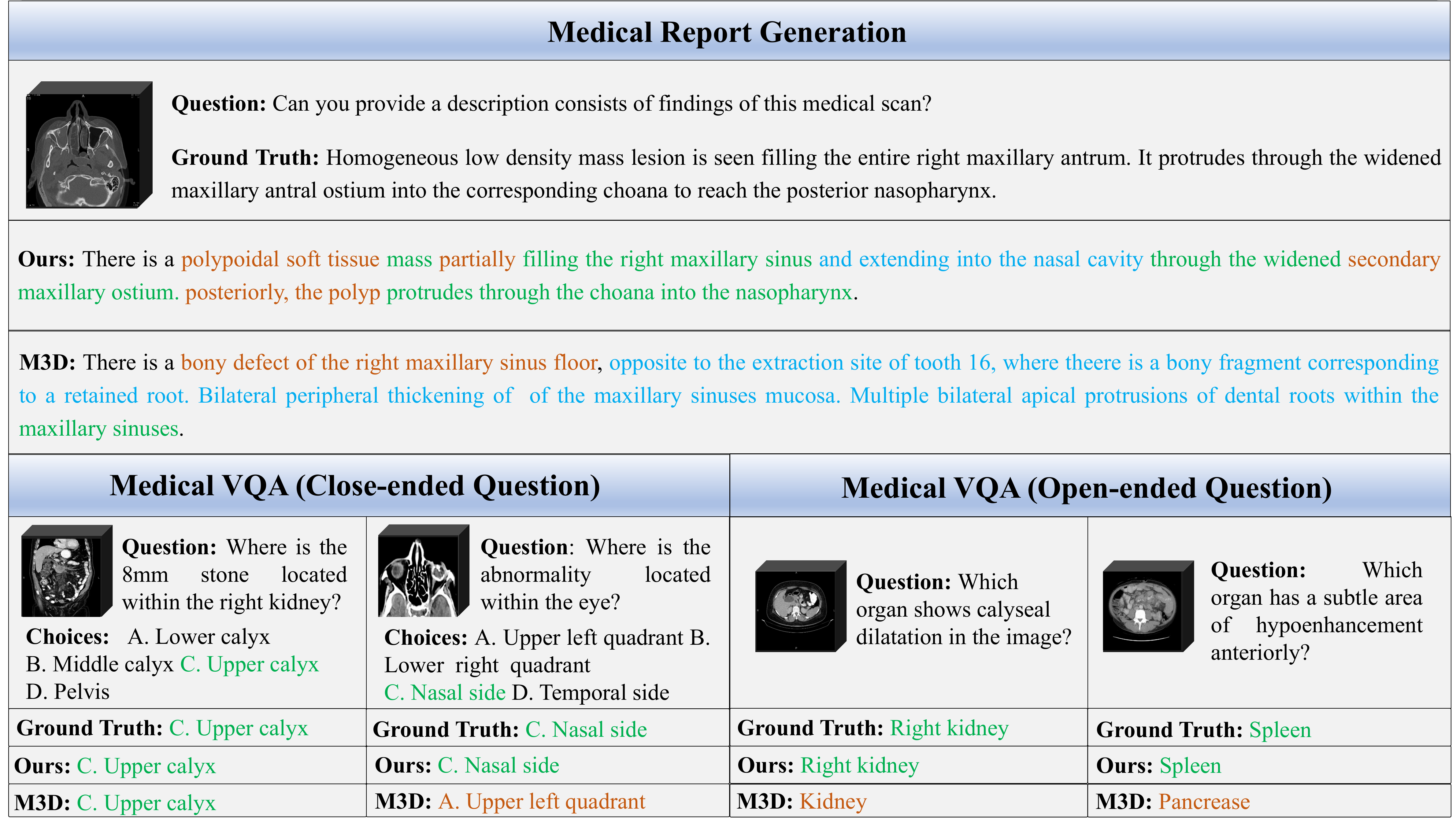}
    \caption{ Qualitative comparisons with our model and ground truth on report generation and visual question answering. Green represents matched text. Orange represents unmatched text. Blue represents excessive text.}
\label{qualitative}
\end{figure*}

\subsubsection{Ablation on Training Strategy} We design a two-stage training strategy to enable the vision encoder to learn task-shared and -specific image features. In Tab.\ref{tab:training-strategy}, we compare our training strategy with other training strategies. 1) only task-shared image features: we only utilize token-level MoE to task-shared image features and omit task-level MoE. 2) only task-specific image features: we directly train TGH-MoE to learn task-specific image features. 3) optimizing LLM with LoRA in Stage 2. As shown in Tab.\ref{tab:training-strategy}, our training strategy outperforms ``Only task-specific". The reason is that our strategy enables the vision encoder to learn task-shared image features in Stage 1 by assigning the two tasks to the same expert. Our training strategy also outperforms "Only task-shared" because it learns task-specific image features in Stage 2 by assigning the two tasks to different experts. Furthermore, we find that freezing the LLM in Stage 2 leads to better results by focusing the training on the task-specific capacity of TGH-MoE.

\subsection{Qualitative Comparisons}
We conduct a qualitative comparison among our method, the SOTA method M3D, and the ground truth. As shown in Fig. \ref{qualitative}, for the report generation task, our model's prediction covers most of the content in the ground truth. For example, our model successfully recognizes the ``haematoma", contour of the ``haematoma", ``mass effect",  and ``generalised cerebral volume loss". These predictions can help the radiologists notice the key problems of the image. However, our model can occasionally make errors in specifying the precise location of a lesion or the severity of a disease.
In contrast, the report generated by M3D is of lower quality and captures only a small portion of the ground truth content.
For the visual question answering task, our method provides correct answers for both closed- and open-ended questions. For example, in response to a question about calyceal dilation, our model correctly specifies the "right kidney." M3D, however, only identifies that a "kidney shows calyceal dilation" but fails to identify the affected kidney.

\section{Conclusion and Future Work}

In this work, we developed a well-trained 2D MLLM for 3D medical images with the strong perceptual capacity of the 2D MLLM inherited. We further propose TGH-MoE to handle both Medical Report Generation (MRG) and Medical Visual Question Answering (MVQA).
Specifically, we first adapt the 2D vision encoder from a pre-trained MLLM to process 3D medical images, thereby leveraging its strong perceptual capacity and image-text alignment. We then propose a Text-Guided Hierarchical MoE (TGH-MoE) that uses the text prompt to dynamically extract task-specific image features for MRG and MVQA. Furthermore, we devise a two-stage training strategy to enable the TGH-MoE to learn both task-shared and -specific features effectively.
Extensive experiments demonstrate that our method outperforms existing state-of-the-art approaches and confirm the effectiveness of each core component in our design.

For future work, we would like to handle task variety from the perspective of the language model of the MLLM. In Tab. \ref{tab:main}, our model is trained on all 106,192 caption samples and all 437,953 VQA samples. In contrast, for the results in Tab. \ref{tab:component}, we use all caption samples but only one-third of the VQA samples. We observe that the performance on report generation improves as the volume of VQA data decreases. This indicates a data imbalance and a seesaw problem stemming from the inherent differences between the tasks.
In this paper, we have identified the variability in image features required for different tasks. However, the linguistic patterns also differ significantly. For instance, MRG requires generating lengthy, descriptive reports, whereas MVQA involves producing concise answers. We leave the challenge of adapting the language model of the MLLM to handle this task variety as a direction for future work.

\bibliographystyle{IEEEtran2}
\bibliography{myref}

\end{document}